\documentclass{svproc}
\usepackage{url}
\usepackage{xcolor}
\usepackage{graphicx}
\usepackage{pifont}
\usepackage{xcolor,pifont}
\usepackage[utf8]{inputenc}

\definecolor{green(ryb)}{rgb}{0.4, 0.69, 0.2}
\definecolor{red(ryb)}{rgb}{1.0, 0.15, 0.07}
\newcommand{\cmark}{\textcolor{green(ryb)}{\ding{51}}}%
\newcommand{\xmark}{\textcolor{red(ryb)}{\ding{55}}}%

\begin{document}
\mainmatter              
\title{The Landscape of Collective Awareness in multi-robot systems}

\titlerunning{Collective Awareness in Multi-Robot Systems.}  
%
\author{Miguel Fernandez-Cortizas\inst{1}, David Perez-Saura\inst{1} \and Ricardo Sanz\inst{2} \and\\ Martin Molina\inst{1,3} \and Pascual Campoy\inst{1}}
\authorrunning{M. Fernandez-Cortizas et al.} 
%

%
\institute{Computer Vision and Aerial Robotics Group (CVAR) \and Autonomous Systems Laboratory (ASL) \and Department of Artificial Intelligence \\Universidad Politécnica de Madrid, Madrid, Spain \email{miguel.fernandez.cortizas@upm.es}}

\maketitle              

\begin{abstract}
The development of collective-aware multi-robot systems is crucial for enhancing the efficiency and robustness of robotic applications in multiple fields. These systems enable collaboration, coordination, and resource sharing among robots, leading to improved scalability, adaptability to dynamic environments, and increased overall system robustness. In this work, we want to provide a brief overview of this research topic and identify open challenges.

\end{abstract}
\section{Introduction}
Multi-robot systems (MRS) are becoming more and more capable day after day. However, there is still a lack of awareness about what is happening around them, that may lead to errors and failures when there are unexpected changes in their operational conditions. Trying to solve this issue,  roboticists work on improving the situational awareness of these robotic systems so that they become more robust and resilient, and they can "know" how to act in order to fulfill their objective coordinately in the best way. In these systems, it is necessary to have not only an "individual awareness" per agent, but also an awareness about the other agents of the system and their interactions, which can be integrated to generate a higher level of awareness, a "Collective Awareness".

In this mapping work, we analyze existing research and literature to identify patterns and knowledge gaps in the field. By examining the current state of the art, the work hopes to provide useful insights into what collective awareness is and how it can be effectively integrated into multi-robot systems to enhance their performance and functionality. In this work, we want to respond to the following questions:

\begin{enumerate}
    \item What \textit{Collective Awareness} is in multi-robot systems? 
    \item How is collective awareness achieved in the literature?
    \item What are the open challenges in the field? 
\end{enumerate}

\section{Literature Review}

In this work, we did a qualitative analysis to evaluate the literature involving the use of collective awareness in multi-robot systems.
To this end, a series of searches were conducted in the Scopus database \footnote{https://www.elsevier.com/products/scopus}. We have searched for "Awareness" AND "robot" adding different keywords such as "collaborative", "distributed", or "collective" to increase the scope of the search up to 517 publications between 1992 and 2024. The number of publications has been increasing since 2007 showing an emerging interest in this research field.

\noindent\textbf{1. Collective Awareness Definition}

In these publications, we found that there is no consensus on what \textit{collective awareness} is and the majority of the works that use this expression do not define it, which makes it difficult to search for a definition. 

Kube et al. \cite{kube1993collective} define Collective Awareness in robotics as ``the ability of multiple robots to work together in a coordinated manner, often using decentralized control''. This approach is inspired by the collective behavior of social insects, which can work together effectively without a centralized leader.

Other authors \cite{salmon2017distributed} use the term "distributed" or "collective" situational awareness as an extension of the former definition of situational awareness \cite{endsley1995toward} as ``the perception of elements in the environment within a volume of time and space, the comprehension of their meaning and the projection of their status in the near future'' but applied to multi-agent systems that perform this comprehension and projection in a collaborative way. 


In works like \cite{Beni1991}, define "Swarm Intelligence" as an approach to the coordination of multiple robots as a system consisting of large numbers of mostly simple physical robots. In this approach, systems operate following the principles of Swarm Robotics, in which the self-organization of a large number of simple (and normally cheap) robots, with only local sensing, communication, and actuation capabilities, generates an emerging behavior that fulfills a concrete task.

\noindent\textbf{2. Collaboration in Multi-robot Systems}

Generally, to achieve some kind of collaboration, each team agent collects information that contributes to its local knowledge of the surrounding environment. To provide a globally consistent view of the entire environment, agents need to share their information with other team members. \cite{berger2023rgs}

In this section, we have summarized some of the most relevant works in this field, trying to identify commonalities and differences between them, leading to the different categorizations that appear in the following. Table \ref{tab:literature} shows a comparison between the most significant works that recapitulate their main characteristics:


\textbf{Data shared}: Depending on the information that is exchanged between agents, three categories can be defined\cite{doherty2021hastily}:
    \textbf{Low-level sensory data}: raw data output from sensors.
    \textbf{Intermediate processed data}: raw data that have been processed to generate some data with a minimal to moderate structure, such as feature/value pairs or the output of sensor fusion algorithms.
    \textbf{Semantic data}: the most abstract layer, in which data are structured around objects with properties and relations, often semantically grounded, including ontological structures, logical structures, or graph-structured knowledge.

\textbf{Type of MRS}: Describes the type of multi-robot system, which determines the behavior and coordination between the agents. There are three different types:
    \textbf{Centralized Systems:} A central unit collecting information from multiple sources. This information is integrated for generating system-level knowledge, which can be used for deciding which is the best actions for each robot. It is effective in complex environments with robots having limited perception, but it lacks robustness, as failure in the central unit results in the entire system failing \cite{chang2023hydramulti}\cite{greve2023collaborative}.
    \textbf{Decentralized Systems:} Multiple robots or agents collaborate without central control. Each agent processes information locally, making decisions based on its observations and interactions with the environment and other agents. Decentralized systems are fault-tolerant, if one agent fails, the system can still operate \cite{RACER2023}\cite{DPPM2023}\cite{fernandez2024multi}.
    \textbf{Swarm Intelligence Systems:} Distributed system composed of self-organizing groups of simple and cost-effective robots. Operating with local sensing, communication, and actuation, they collectively generate emergent behavior\cite{Potts-2019} to accomplish specific tasks \cite{jurt2022collective}.

\textbf{System state awareness: } A binary property of the MRS that describes if the system generates a global view of the state of the system.

\textbf{Mission awareness:} A binary property of the MRS that describes if the cognitive agents of the system know what is the mission that they are performing so that they can generate plans according to it.

\begin{table}[!htb]
    \centering
    \vspace{-0.3cm}
    \begin{tabular}{c|c|c|c|c|c}
         \textbf{Work} & \textbf{Task} & \textbf{Data Shared} & \textbf{Type of MRS} & \textbf{SSA} & \textbf{MA} \\  
         \hline
         \cite{meng2008distributed} & Construction & L & Swarm & \xmark & \cmark \\
         \hline
         \cite{ebert2022hybrid} & Exploration & L & Swarm & \xmark & \cmark \\
         \hline
         \cite{chang2023hydramulti}\cite{greve2023collaborative} & SLAM & L,S & Centralized & \cmark & \xmark  \\
         \hline
         \cite{luna2023spiral} & Area coverage & L & Centralized & \cmark & \xmark \\
         \hline
         \cite{jurt2022collective} & Manipulation & S & Decentralized & \cmark & \cmark \\
         \hline
         \cite{RACER2023}\cite{DPPM2023} & Exploration & L & Decentralized & \cmark & \cmark \\
         \hline
         \cite{doherty2021hastily} & Exploration & I, L, S & Decentralized& \cmark & \cmark \\
         \hline
         \cite{fernandez2024multi} & SLAM & L,S & Decentralized & \cmark & \xmark \\ 
         \hline
    \end{tabular}
    \vspace{0.3cm}
    \caption{Comparison between some of the most relevant works. Data shared is categorized as S: Semantic data, I: Intermediate data, L: Low-level sensory data. SSA: System State Awareness. MA: Mission Awareness. }
    \vspace{-0.6cm}
    \label{tab:literature}
\end{table}

\section{Conclusions}

In this study we explore the field of collective awareness in multi-robot applications, finding that although there is not a lot of work in this field, there are promising results that can improve the capabilities of MRS. Below we collect our insights of this mini-review when trying to answer the former questions that were formulated at the start of this work.

\subsubsection{What is Collective Awareness?}

From the previous categorization we have found that there are multiple ways of achieving collaboration in Multi-Robot Systems; however, we consider that not all the Multi-agent systems in which there is some kind of collaboration have this "Collective Awareness". Based on the above, we propose the following definition.

\begin{definition}
A multi-agent system has Collective Awareness when: 
\begin{enumerate}
    \item Each agent has some degree of awareness about itself, about the environment that surrounds it, and about what is its goal.
    \item Different agents are able to communicate between them.
    \item The system as a whole is able to integrate all the information provided by the agents to extract knowledge that can be used to improve the knowledge of the system or the capabilities of the system.
\end{enumerate}
\end{definition}

We consider that the Collective Awareness extends the sum of the individual Situational Awareness of each agent, since new observations between agents can be taken into account to extend the Individual Awareness of each robot or to generate emergent knowledge about the status of the system, which can lead into improved emergent behavior.

Some of the systems previously described do not fit in this definition, like the swarm systems, because each agent only has a very restricted knowledge of the whole system. In that case, the system only has a partial collective awareness, since it does not have awareness of the complete collective.  
\vspace{-0.3cm}
\subsubsection{How is collective awareness achieved in the literature?}

In the literature, we can observe that the awareness of the robots is achieved by generating models of interest of the environment or the robot itself and sharing them with the rest of the system. There is a trend to take advantage of the use of high-level semantic knowledge, what can improve the performance and robustness of multi-robot systems as is demonstrated in \cite{fernandez2024multi}\cite{chang2023hydramulti}\cite{doherty2021hastily}.

\vspace{-0.3cm}
\subsubsection{What are the open challenges in the field?}

One of the main limitations of these approaches is the lack of generality; most systems are tailored specifically for a very concrete use case, which limits the expansion or reutilization of these systems \cite{kernbach2011awareness}. Some works provide tools for extending the knowledge of collective-aware systems with the use of formalizations such as ontologies. However, these systems does not take into account relevant aspects such as knowledge uncertainty, risk associated with one decision, or failure tolerance \cite{doherty2021hastily}. Moreover, in the majority of systems, a secure and stable communication link between robots is taken as guaranteed, which may not be present in multiple applications.

\vspace{-0.2cm}
\section*{Aknowledgments}
\small This work was funded by: European Union's Horizon Europe Project No. 101070254 CORESENSE, project COPILOT ref. 2020/EMT6368, funded by the Madrid Government under the R\&D Synergic Projects Program, project INSERTION ref. ID2021-127648OBC32 funded by the Spanish Ministry of Science and Innovation. The work of the second author is supported by the Spanish Ministry of Science and Innovation under its Program for Technical Assistants PTA2021-020671.
\bibliographystyle{bibtex/spmpsci.bst}
\bibliography{references.bib}
\end{document}